\documentclass[12pt,a4paper]{article}

\usepackage[british]{babel}

\usepackage[a4paper,top=2cm,bottom=2cm,left=2.5cm,right=2.5cm,marginparwidth=1.75cm]{geometry}

\usepackage{amsmath}
\usepackage{graphicx}
\usepackage[colorlinks=true, allcolors=blue]{hyperref}
\usepackage[title]{appendix}
\usepackage{mathrsfs}
\usepackage{amsfonts}
\usepackage{booktabs} 
\usepackage{caption}  
\usepackage{threeparttable} 
\usepackage{algorithm}
\usepackage{algorithmicx}
\usepackage{algpseudocode}
\usepackage{listings}
\usepackage{enumitem}
\usepackage{chngcntr}
\usepackage{booktabs}
\usepackage{lipsum}
\usepackage{subcaption}
\usepackage{authblk}
\usepackage[T1]{fontenc}    
\usepackage{csquotes}       
\usepackage{diagbox}

\usepackage{tikz, amsmath, amssymb, tikz-layers}
\usetikzlibrary{intersections, calc, positioning, arrows, matrix}

\usepackage{setspace}
\usepackage{titlesec}
\titleformat{\section} 
{\normalfont\Large\bfseries}{\thesection.}{1em}{}

\usepackage{lineno} 

\rightlinenumbers 

\usepackage{float}   
\usepackage{caption} 
\usepackage{comment} 



\title{Discounted Pseudocosts in MILP}

\author[1,*]{Krunal Kishor Patel}

\affil[1]{\small CERC, Polytechnique Montr\'eal, 2500 Chemin de Polytechnique, Montr\'eal, H3T 1J4, QC, Canada}
\affil[*]{Corresponding author: \texttt{krunal.patel@polymtl.ca}}

\date{}  

\begin{document}
	\maketitle

\begin{abstract}
	In this article, we introduce the concept of discounted pseudocosts, inspired by discounted total reward in reinforcement learning, and explore their application in mixed-integer linear programming (MILP). Traditional pseudocosts estimate changes in the objective function due to variable bound changes during the branch-and-bound process. By integrating reinforcement learning concepts, we propose a novel approach incorporating a forward-looking perspective into pseudocost estimation. We present the motivation behind discounted pseudocosts and discuss how they represent the anticipated reward for branching after one level of exploration in the MILP problem space. Initial experiments on MIPLIB 2017 benchmark instances demonstrate the potential of discounted pseudocosts to enhance branching strategies and accelerate the solution process for challenging MILP problems.
\end{abstract}

	

	\section*{Acknowledgements}
	I would like to thank the SCIP developers at ZIB, especially Suresh Bolusani, Ksenia Bestuzheva, Leon Eifler, Ambros Gleixner, Alexander Hoen, Dominik Kamp, Julian Manns, Gioni Mexi, and Mark Turner, for their assistance with understanding the SCIP code and helping with benchmarking and testing. Special thanks to Timo Berthold for suggesting extensions to this project. I also thank Paul Strang for helping me find articles for literature review. Finally, I thank CERC and ZIB for funding my visit to Berlin to work on this project.

	\section{Introduction}
	
	\subsection{Reinforcement Learning basics}
	
	In reinforcement learning (RL), an agent operates within an environment. At any given timestamp, the environment specifies its state. The agent has a set of actions available to choose from. The agent selects some action. This modifies the current state of the environment and the agent gets some reward from the environment.
	
	The primary objective of the agent is to maximize the total reward obtained through its actions. The reward that the agent receives immediately after any action is called an ‘immediate reward’. The agent sometimes can pick a greedy action that maximizes the immediate reward without caring about the future states or rewards. But, mostly we are interested in the total reward. In reinforcement learning and in real life as well, we care more about the immediate reward than the future rewards. So, we use discounted total reward (DTR) by multiplying the future rewards by a constant factor called the discount factor ($\gamma$). See equation (\ref{dtr}).
	
	\begin{equation}
		\label{dtr}
		DTR = R_0 + \gamma R_1 + \gamma^2 R_2 + \dots
	\end{equation}

	\subsection{Related work}
	
	Recently, several researchers have applied RL to Mixed Integer Linear Programming (MILP). For this article, we focus on the branching part. We list some of the work that uses RL for MILP. This list is not complete.
	
	\begin{itemize}
		\item Reinforcement learning for variable selection in a branch and bound algorithm \cite{etheve2020reinforcement} 
		\item Branch Ranking for Efficient Mixed-Integer Programming via Offline Ranking-Based Policy Learning \cite{huang2022branch} 
		\item Reinforcement learning for branch-and-bound optimisation using retrospective trajectories \cite{parsonson2023reinforcement} 
		\item An improved reinforcement learning algorithm for learning to branch \cite{qu2022improved}
		\item Learning to branch with tree MDPs \cite{scavuzzo2022learning}
		\item TreeDQN: Learning to minimize Branch-and-Bound tree \cite{sorokin2023treedqn}
		\item Improving learning to branch via reinforcement learning \cite{sun2020improving}
		\item Deep reinforcement learning for exact combinatorial optimization: Learning to branch \cite{zhang2022deep}
	\end{itemize}

	One of the common issues with all these works is that they work on specific instance types (see \cite{gasse2019exact}) and not on general MILPs. However, these work show very good results in those specific instances. They also require some offline training. We attempt to address these challenges in this work.
	
	
	\section{Discounted pseudocosts}

	When using RL for MILP, the objective is to design an effective reward function. One example of a reward function for branching is to give ‘-1’ as a reward for each node. The agent is incentivized to minimize the number of nodes in the search tree. Some of the previous work cited above used this reward function.

	We can also see pseudocosts \cite{benichou1971pscost} as a reward function for branching on a particular variable. Pseudocosts capture the improvement in relaxation value for just one level. So, this is equivalent to the concept of ‘immediate reward’ in RL. 

	Our objective is to extend pseudocosts to capture the linear programming (LP) relaxation bound improvements observed later in the tree, thereby aligning more closely with the concept of ‘total reward’ or ‘discounted total reward’ in RL.

	We refer to the modified pseudocosts as discounted pseudocosts. Assume we are at a node with the LP relaxation objective value $L$. Now we branch on a variable $x$ and on the branch $x \leq \lfloor x \rfloor$, we get the LP relaxation objective value $L_x$. Our traditional pseudocost ($PS_0(x)$) update for $x$ would be $\dfrac{L_x - L}{\{x\}}$, where $\{x\}$ denotes the fractional part of variable $x$ in the relaxation solution. Now let us assume we branch here on variable $y$, and on the branch $y \leq \lfloor y \rfloor$ we get the LP relaxation objective value $L_y$. Here we update the traditional pseudocost of variable $y$ ($PS_0(y)$) by adding datapoint $\dfrac{L_y - L_x}{\{y\}}$. 

	In discounted pseudocosts, we additionally maintain the first level pseudocosts for each variable. For the case described above, we update the first level discounted pseudocost of $x$ ($PS_1(x)$) by adding datapoint $\dfrac{L_y - L_x}{\{x\}}$. In other words, we give some credit of second LP relaxation objective gain to variable $x$. This new update is for the first level which is discounted by a factor gamma. Further levels are similarly discounted by higher powers of gamma. The complete discounted pseudocost used for branching is described in equation (\ref{dpscost}).
	
	\begin{equation}
		\label{dpscost}
		DPS(x) = PS_0(x) + \gamma PS_1(x) + \gamma^2 PS_2(x) + \dots
	\end{equation}

	For the first implementation, we used one-level discounted pseudocosts to make the implementation simple and learn what a good discount factor should be.
	
	\sloppy
	There is also another way of looking at discounted pseudocosts. Traditional pseudocosts approximate the strong branching choice whereas one-level discounted pseudocosts approximate the lookahead branching \cite{glankwamdee2006lookahead} choice (when $\gamma = 1$). The lookahead branching gives smaller trees compared to strong branching, however, we do not use it because it requires a lot more computational effort.
	
	Extending discounted pseudocosts concept to reliability branching or hybrid branching \cite{achterberg2005reliability} is one major challenge we faced. In reliability branching we use strong branching for variables when the pseudocosts are not reliable. \cite{hendel2015pscostvariance} describes various ways to measure reliability. In discounted pseudocosts we have two levels of pseudocosts (level 0 and level 1). So making them reliable is even harder. 
	
	We observed that using the discounted pseudocosts when either level is not reliable degrades the performance. We switch to reliability branching using regular pseudocosts ($PS_0$) for evaluating branching candidates when at least one candidate variable has either of the two levels unreliable. This way very few instances are affected but solves the performance degradation to some extent.

	\section{Computational results}

	We use the MIPLIB dataset and solve instances with SCIP version 9.0 \cite{bolusani2024scipoptimizationsuite90} using three different seeds. This gives us a total of 720 instance-seed pairs. Although we have experimented with turning off presolvers, separators, and heuristics, in this article, we only present the results with the default setting where only the branching rule is changed. The results for the other experiments were similar to the ones presented below.
	
	First, we compare the results with pseudocosts vs discounted pseudocosts in Table \ref{tab:pscostdefault}. The columns `pscost' represent the pseudocost branching rule and the `dpscost' represent one-level discounted pseudocost branching rule. We present the number of solved instances, shifted geometric mean (shifted by 1s) of solving time (ratio for the `dpscost' variant), and shifted geometric mean (shifted by 100 nodes) of number of nodes (ratio for the `dpscost' variant). The brackets show results for all as well as affected instances. The rows $\geq Xs$ show the results for instances in which at least one variant took more than $X$ seconds to solve. We used a time limit of two hours for this experiment.
	
	There are three more instances solved and nearly no significant improvement in either solving times or nodes. We also performed similar experiments with all features turned off and solution cutoff provided for which we observed similar results (not presented in this article). These numbers show a minor improvement which can also be called noise. The last bracket ($\geq 1000s$) shows 8\% improvement in solving time.
	
	The choice of discount factor was also very hard to make because of such benchmark results. But we settled on 0.2 after a few experiments. We expect the optimal value of the discount factor to lie in the $[0.2, 0.5]$ interval.
	
	
	Next, we show the comparison with reliability branching in Table \ref{tab:rpscostMIPLIBdefault}. The time limit was one hour for this experiment. The numbers show similar improvement as in the comparison with pseudocosts. Here we see that fewer instances are affected. The variant with discounted pseudocosts (rdpscost) solves 3 more instances than the default variant (rpscost). We observe no significant improvement in time or node except for the last bracket ($\geq 1000s$).
	
	Finally, since SCIP is also a MINLP solver, we also run the benchmarks for MINLP. The results are presented in Table \ref{tab:rpscostminlpdefault}. We used a time limit of one hour for this experiment to evaluate a total of 507 instance seed combinations. We observe no major impact as expected, but primal-dual integral (PDI) was surprisingly improved and 3 more instances were solved by the discounted pseudocost variant (rdpscost) in the last bracket ($\geq 1000s$)..
	
	Overall, the results are not groundbreaking. The improvements are very small and can be called noise. The only reason we believe this is not noise is that similar results are obtained in many comparisons. The implementation is added to SCIP and turned off by default until we observe gains that are beyond noise margins.
	
\begin{table}[H]
	\centering
	\caption{Comparison with pseudocost branching rule.}
	\label{tab:pscostdefault}
	\begin{tabular}{lrrrrrrr}
		\toprule
		& & \multicolumn{2}{c}{Solved} &\multicolumn{2}{c}{Time(1)}&\multicolumn{2}{c}{Node(100)} \\
		\cmidrule(lr){3-4} \cmidrule(lr){5-6} \cmidrule(lr){7-8}  
		Bracket & \multicolumn{1}{c}{Count} & \multicolumn{1}{c}{pscost} & \multicolumn{1}{c}{dpscost}& \multicolumn{1}{c}{pscost} & \multicolumn{1}{c}{dpscost} & \multicolumn{1}{c}{pscost} & \multicolumn{1}{c}{dpscost}\\
		\midrule
		All & 720 & 323 & 326 & 1376.06 & 0.99 & 6911 & 1.01 \\
		Affected & 258 & 244 & 247 & 418.17 & 0.97 & 9317 & 1.01 \\
		$\geq$ 0s & 337 & 323 & 326 & 215.39 & 0.98 & 3246 & 1.01 \\
		$\geq$ 1s & 334 & 320 & 323 & 225.27 & 0.98 & 3353 & 1.01 \\
		$\geq$ 10s & 300 & 286 & 289 & 341.09 & 0.98 & 4746 & 1.01 \\
		$\geq$ 100s & 207 & 193 & 196 & 978.76 & 0.96 & 13477 & 0.99 \\
		$\geq$ 1000s & 118 & 104 & 107 & 2405.05 & 0.92 & 32508 & 0.96 \\
		\bottomrule
	\end{tabular}
\end{table}

\begin{table}[H]
	\centering
	\caption{Comparison with reliable pseudocost branching rule.}
	\label{tab:rpscostMIPLIBdefault}
	\begin{tabular}{lrrrrrrr}
		\toprule
		& & \multicolumn{2}{c}{Solved} &\multicolumn{2}{c}{Time(1)}&\multicolumn{2}{c}{Node(100)} \\
		\cmidrule(lr){3-4} \cmidrule(lr){5-6} \cmidrule(lr){7-8}  
		Bracket & \multicolumn{1}{c}{Count} & \multicolumn{1}{c}{rpscost} & \multicolumn{1}{c}{rdpscost}& \multicolumn{1}{c}{rpscost} & \multicolumn{1}{c}{rdpscost} & \multicolumn{1}{c}{rpscost} & \multicolumn{1}{c}{rdpscost}\\
		\midrule
			All & 720 & 345 & 348 & 811.19 & 1.00 & 3158 & 1.01 \\
			Affected & 158 & 151 & 154 & 541.87 & 0.99 & 11742 & 1.00 \\
			$\geq$ 0s & 352 & 345 & 348 & 171.65 & 1.00 & 1896 & 1.02 \\
			$\geq$ 1s & 349 & 342 & 345 & 178.84 & 1.00 & 1948 & 1.02 \\
			$\geq$ 10s & 310 & 303 & 306 & 278.4 & 1.00 & 2701 & 1.02 \\
			$\geq$ 100s & 219 & 212 & 215 & 685.25 & 1.00 & 6995 & 1.03 \\
			$\geq$ 1000s & 94 & 87 & 90 & 1853.23 & 0.94 & 19145 & 1.02 \\
			\bottomrule
		\end{tabular}
	\end{table}

\begin{table}[H]
	\centering
	\caption{Comparison with reliable pseudocost branching rule on MINLP dataset.}
	\label{tab:rpscostminlpdefault}
	\begin{tabular}{lrrrrrrrrr}
		\toprule
		& & \multicolumn{2}{c}{Solved} &\multicolumn{2}{c}{Time(1)}&\multicolumn{2}{c}{Node(100)} &\multicolumn{2}{c}{PDI(100)} \\
		\cmidrule(lr){3-4} \cmidrule(lr){5-6} \cmidrule(lr){7-8}  \cmidrule(lr){9-10}
		Bracket & \multicolumn{1}{c}{Count} & \multicolumn{1}{c}{rpscost} & \multicolumn{1}{c}{rdpscost}& \multicolumn{1}{c}{rpscost} & \multicolumn{1}{c}{rdpscost} & \multicolumn{1}{c}{rpscost} & \multicolumn{1}{c}{rdpscost} & \multicolumn{1}{c}{rpscost} & \multicolumn{1}{c}{rdpscost}\\
		\midrule
		All & 507 & 466 & 469 & 27.41 & 1.00 & 2420 & 1.00 & 717.5 & 0.99 \\
		Affected & 129 & 125 & 128 & 81.03 & 1.00 & 11849 & 0.99 & 1683.9 & 0.98 \\
		$\geq$ 0s & 470 & 466 & 469 & 21.88 & 1.00 & 2000 & 1.00 & 576 & 0.99 \\
		$\geq$ 1s & 430 & 426 & 429 & 28.31 & 1.00 & 2643 & 1.00 & 694.5 & 0.99 \\
		$\geq$ 10s & 275 & 271 & 274 & 79.19 & 0.99 & 5260 & 1.00 & 1538.5 & 0.98 \\
		$\geq$ 100s & 109 & 105 & 108 & 377.48 & 1.00 & 31079 & 1.01 & 4295.6 & 0.98 \\
		$\geq$ 1000s & 27 & 23 & 26 & 1875.48 & 1.00 & 98292 & 1.01 & 21738.1 & 0.92 \\
		\bottomrule
	\end{tabular}
\end{table}

	\section{Future work and Conclusion}

	There are many possible extensions of discounted pseudocosts to improve the current results. One idea is to replace pseudocosts with discounted pseudocosts in the primal heuristics that use them.
	
	The natural extension of going for two or more levels seems a bit far as of now. The major issue is with the reliability of higher-level discounted pseudocosts.
	
	The following ideas were suggested by Timo Berthold.
	
	\begin{itemize}
		\item Only update the first level of discounted pseudocost if the first branched variable ($x$ in the example) is close to the second branched variable ($y$ in the example) in the variable constraint graph or in general are known to be related. This should hopefully decrease noise in the updates and give better results.
		\item We can also use discounting to extend conflict score or other fields that we maintain in the variable history and use for making the branching decision.
	\end{itemize}

	In conclusion, we presented an idea of discounted pseudocosts where we can use RL concepts in MIP for branching. Discounted pseudocosts do not require any offline training and therefore work on general MILPs. The initial implementation with very little tuning for the discount factor shows a small improvement in the solving times for harder instances.

	\renewcommand\bibname{REFERENCES}
	\bibliography{references}
	\bibliographystyle{apalike}			
	
	

	
	\renewcommand\theequation{\Alph{section}\arabic{equation}} 
	\counterwithin*{equation}{section} 
	\renewcommand\thefigure{\Alph{section}\arabic{figure}} 
	\counterwithin*{figure}{section} 
	\renewcommand\thetable{\Alph{section}\arabic{table}} 
	\counterwithin*{table}{section} 

\end{document}